# Evaluating the Navigation Capabilities of a Modified COAST Guidewire Robot in an Anatomical Phantom Model

Timothy A. Brumfiel[1*†], Revanth Konda[1*], Drew Elliott[1], and Jaydev P. Desai[1]

[1] *Medical Robotics and Automation (Robomed) Laboratory, Georgia Institute of Technology*

*Equal Contribution, † Corresponding Author, Email: tbrumfiel3@gatech.edu

## INTRODUCTION

To address the issues that arise due to the manual navigation of guidewires in endovascular interventions, research in medical robotics has taken a strong interest in developing robotically steerable guidewires, which offer the possibility of enhanced maneuverability and navigation, as the tip of the guidewire can be actively steered [1], [2]. The COaxially Aligned STeerable (COAST) guidewire robot has the ability to generate a wide variety of motions including bending motion with different bending lengths, follow-the-leader motion, and feedforward motion. In our past studies, we have explored different designs of the COAST guidewire robot [2], [3] and developed modeling, control, and sensing strategies for the COAST guidewire robot [4], [5]. In this study, the performance of a modified COAST guidewire robot is evaluated by conducting navigation experiments in an anatomical phantom model with pulsatile flow. The modified COAST guidewire robot is a simplified version of the COAST guidewire robot and consists of two tubes as opposed to three tubes. Through this study, we demonstrate the effectiveness of the modified COAST guidewire robot in navigating the tortuous phantom vasculature.

## MATERIALS AND METHODS

The modified COAST guidewire robot consists of two nitinol tubes. The outer tube, with an inner diameter (ID) and outer diameter (OD) of 0.30 mm and 0.35 mm, respectively, consists of two machined segments. The distal segment is laser micromachined with a unidirectional asymmetric notch (UAN) pattern, to form a bending joint [2], which is followed by another machined segment consisting of a sparse trapezoidal notch pattern. This segment with passive notches, facilitates gradual increase in stiffness along the length and prevents the outer tube from kinking at the base of the bending joint while being advanced into distally curved vessels. The tip of the outer tube is attached with a nitinol tendon (0.076 mm OD), using steel-reinforced epoxy. The inner tube, with an ID and OD of 0.10 mm and 0.20 mm, respectively, is an unmachined tube and is used to vary the bending length at the tip [2]. The outer tube was covered by a 35D durometer Pebax® sheath (ID: 0.51 mm OD: 0.66 mm) to prevent backflow of water into the guidewire.

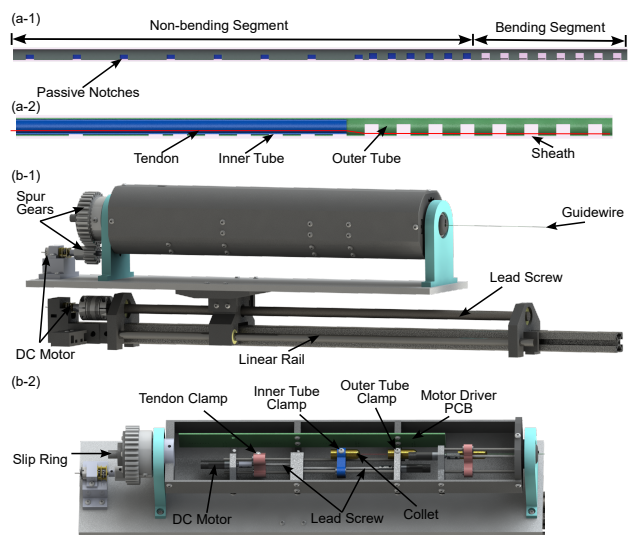

Fig. 1 A rendering of (a-1) the combined structure with the two tubes and (a-2) the cross-section of the modified COAST guidewire robot. (b-1) Perspective view and (b-2) top view of the actuation mechanism.

The guidewire is attached to an actuation mechanism (Fig. 1(b-1)) consisting of a 3D-printed capsule with four modules, three of which were used to attach the tubes and the remaining one module was used to attach the tendon. Since a modified COAST guidewire was used in this study, only three modules comprising two tubes and a tendon were utilized. As shown in Fig. 1(b-2), a 3D-printed bushing is attached to the proximal end of the outer tube using epoxy. This bushing was then placed in the collet of the corresponding module and secured. A similar strategy is used to attach the inner tube to its respective module, which consists of a motorized linear rail for insertion and retraction. The tendon attachment module also consists of a motorized linear rail with the tendon secured using a screw, washer, and nut. A spur gear is attached to the end of the actuation mechanism to enable roll motion of the guidewire. The assembly is placed on a motorized linear rail for insertion and retraction of the guidewire into the phantom model. The motors of the actuation mechanism were teleoperated using a wireless controller (Xbox 360 Controller, Microsoft,WA).

As shown in Fig. 2(a-1), the experimental setup consists of a phantom model connected to a linear actuating pump with viscoelastic impedance (Vivitro Labs Inc., London, Ontario, Canada). The phantom model was placed under a C-arm and covered with a black cloth to simulate a clinical setting, where visual feedback is obtained only from the C-arm fluoroscopic imaging. As shown in Fig. 2(a-2), a phantom model, depicting an aortic bifurcation with different branches, was used.

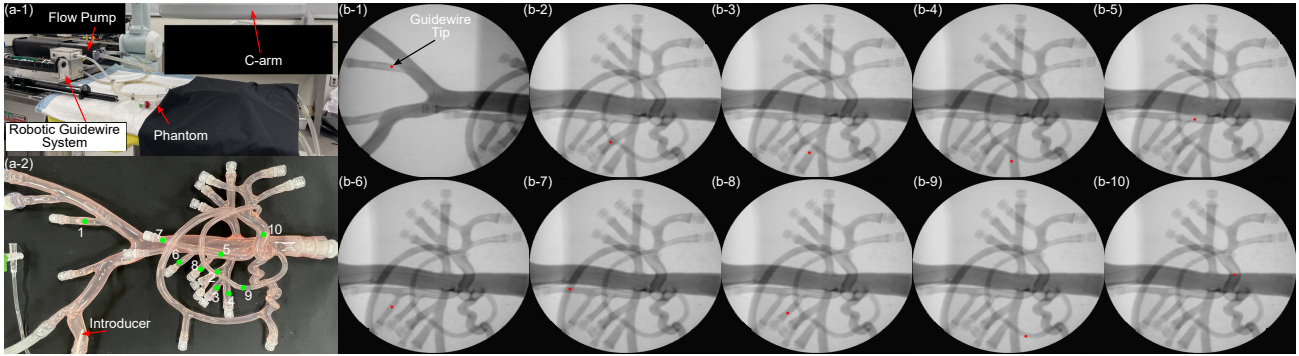

Fig. 2 Image depicting (a-1) the experimental setup and (a-2) the phantom model used in this study. (b-1)-(b-10) Fluoroscopic images depicting traversal of the guidewire into different locations within the phantom model.

## RESULTS

First, the guidewire was teleoperated to enter the internal iliac artery after traversing the aortic bifurcation (Fig. 2(b-1)). Next, the guidewire was traversed up the descending aorta and navigated into the three branches of the left renal artery (Fig. 2(b-2)–(b-4)). This was followed by traversing the superior mesenteric artery. The guidewire was rotated until the tip of the guidewire was approximately oriented toward the opening of the superior mesenteric artery. The guidewire was then navigated by simultaneously actuating the bending joint, increasing the bending length of the joint, and advancing the guidewire. After reaching the location shown in Fig. 2(b-5), the bending length was decreased while relaxing the bending joint. The end of this vessel consisted of three branches. The first branch was traversed by appropriately orienting the guidewire tip with the opening of the branch, followed by simultaneously actuating the bending joint, increasing the bending length of the joint, and advancing the guidewire, as shown in Fig. 2(b-6). Similar procedure was followed to traverse the remaining two branches as shown in Fig. 2(b-7) and Fig. 2(b-8).

This was followed by retracting the guidewire into the aorta to traverse the different branches of the celiac trunk. The celiac trunk was traversed by appropriately orienting the guidewire tip with the opening of the branch, followed by simultaneously actuating the bending joint, increasing the bending length of the joint, and advancing the guidewire. This vessel consisted of three branches namely common hepatic artery, left gastric artery, and splenic artery. First, the left gastric artery was traversed (Fig. 2(b-9)) and then the guidewire was navigated to the common hepatic artery (Fig. 2(b-10)). For both cases, the procedure described to navigate the previous set of vessels was followed. It is noted that during the experiments, the insertion and retraction of the guidewire in the phantom was assisted by hand as the length of the guidewire between the entry point of the phantom and the proximal end of the actuation mechanism experienced buckling.

## CONCLUSIONS AND DISCUSSION

In this article, the navigation capabilities of a modified COAST guidewire robot were qualitatively evaluated in a phantom model equipped with a pulsatile flow system. The guidewire was teleoperated to traverse multiple locations within the aortic bifurcation-based anatomical phantom. Overall, the performance of the guidewire was satisfactory with no major issues encountered during teleoperation. Future work will focus on the following aspects. Firstly, the radiopacity of the guidewire will be enhanced. Secondly, the torque transmission of the guidewire system will be improved. In the current actuation mechanism, the transmission of torque from the proximal end of the guidewire to the distal end is not efficient, leading to a lag in rotation at the distal end of the guidewire.

## ACKNOWLEDGEMENT

Research reported in this publication was supported in part by the National Heart, Lung, And Blood Institute of the National Institutes of Health under Award Number R01HL144714. The content is solely the responsibility of the authors and does not necessarily represent the official views of the National Institutes of Health.

## REFERENCES

[1] Y. Kim, E. Genevriere, P. Harker, J. Choe, M. Balicki, R. W. Regenhardt, J. E. Vranic, A. A. Dmytriw, A. B. Patel, and X. Zhao, "Telerobotic neurovascular interventions with magnetic manipulation," *Science Robotics*, vol. 7, no. 65, p. eabg9907, 2022.
[2] Y. Chitalia, A. Sarma, T. A. Brumfiel, N. J. Deaton, M. Sheft, and J. P. Desai, "Model-based design of the COAST guidewire robot for large deflection," *IEEE Robotics and Automation Letters*, vol. 8, no. 9, pp. 5345–5352, 2023.
[3] S. R. Ravigopal, R. Konda, N. Malhotra, and J. P. Desai, "Design, analysis, and demonstration of the coast guidewire robot with middle tube rotation for endovascular interventions," *Scientific Reports*, vol. 14, p. 27629, 2024.
[4] A. Sarma, T. A. Brumfiel, Y. Chitalia, and J. P. Desai, "Kinematic modeling and jacobian-based control of the COAST guidewire robot," *IEEE Transactions on Medical Robotics and Bionics*, vol. 4, no. 4, pp. 967–975, 2022.
[5] N. J. Deaton, T. A. Brumfiel, A. Sarma, and J. P. Desai, "Simultaneous shape and tip force sensing for the COAST guidewire robot," *IEEE Robotics and Automation Letters*, vol. 8, no. 6, pp. 3725–3731, 2023.